\documentclass[10pt,twocolumn,letterpaper]{article}
\usepackage{cvpr}      
\usepackage[utf8]{inputenc} 
\usepackage[T1]{fontenc}    
\usepackage{hyperref}       
\usepackage{url}            
\usepackage{booktabs}       
\usepackage{amsfonts}       
\usepackage{nicefrac}       
\usepackage{microtype}
\usepackage{wrapfig}

\usepackage{xcolor} 
\usepackage{color}
\usepackage{amsmath}
\usepackage{lipsum}
\usepackage{todonotes}
\usepackage{multicol}
\usepackage{enumitem}
\usepackage{graphicx}
\usepackage{balance}  
\usepackage{amsmath}
\usepackage[linesnumbered,ruled,vlined]{algorithm2e}
\usepackage{algpseudocode}
\usepackage{latexsym}
\usepackage{multirow}
\usepackage{multicol}
\usepackage{color}
\usepackage{dashrule}
\usepackage{extarrows}
\usepackage{dsfont}
\usepackage{centernot}
\usepackage{hyperref}
\usepackage{natbib}
\usepackage[all,cmtip]{xy} 
\usepackage{xcolor}
\usepackage{colortbl}

\definecolor{nav}{RGB}{0,0,128}

\definecolor{mygray}{gray}{0.7}

\newcommand{\model}{\textsc{DiffuSSM}\xspace}
\definecolor{mylightgray}{rgb}{0.9,0.9,0.9} 
\definecolor{cvprblue}{rgb}{0.21,0.49,0.74}
\def\paperID{8684} 
\def\confName{CVPR}
\def\confYear{2024}

\title{Diffusion Models Without Attention}

\author{Jing Nathan Yan$^{1}$$^*$,~~Jiatao Gu$^{2}$$^*$,~~Alexander M. Rush$^{1}$ \\
$^{1}$Cornell University, $^{2}$Apple \\
\texttt{\{jy858, arush\}@cornell.edu, jgu32@apple.com} }

\usepackage{fancyhdr} 
\renewcommand{\headrulewidth}{0pt}
\begin{document}

\twocolumn[{
\renewcommand\twocolumn[1][]{#1}
\maketitle
    \vspace{-2em}
    \includegraphics[width=1\linewidth]{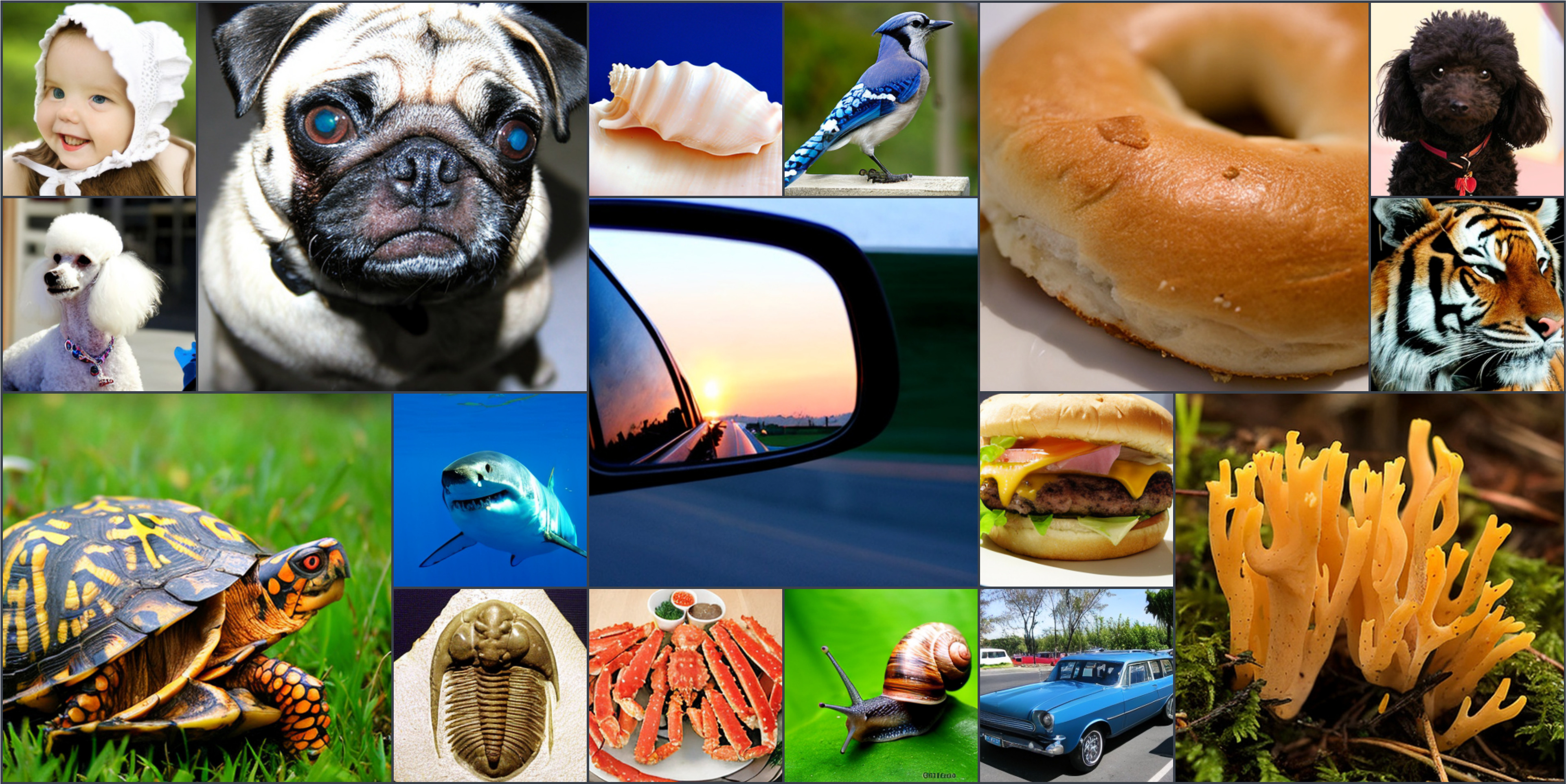}
     \captionof{figure}{Selected samples generated by class-conditional \model trained on ImageNet 256 $\times$ 256 and 512 $\times$ 512 resolutions. }\label{fig:teaser}
     \vspace{25pt}
}]

\renewcommand{\thefootnote}{*}
\footnotetext{Equal contribution.}
\renewcommand{\thefootnote}{\arabic{footnote}}

\begin{abstract}
\begin{figure*}[t]
    \centering    
    \includegraphics[width=1\linewidth]{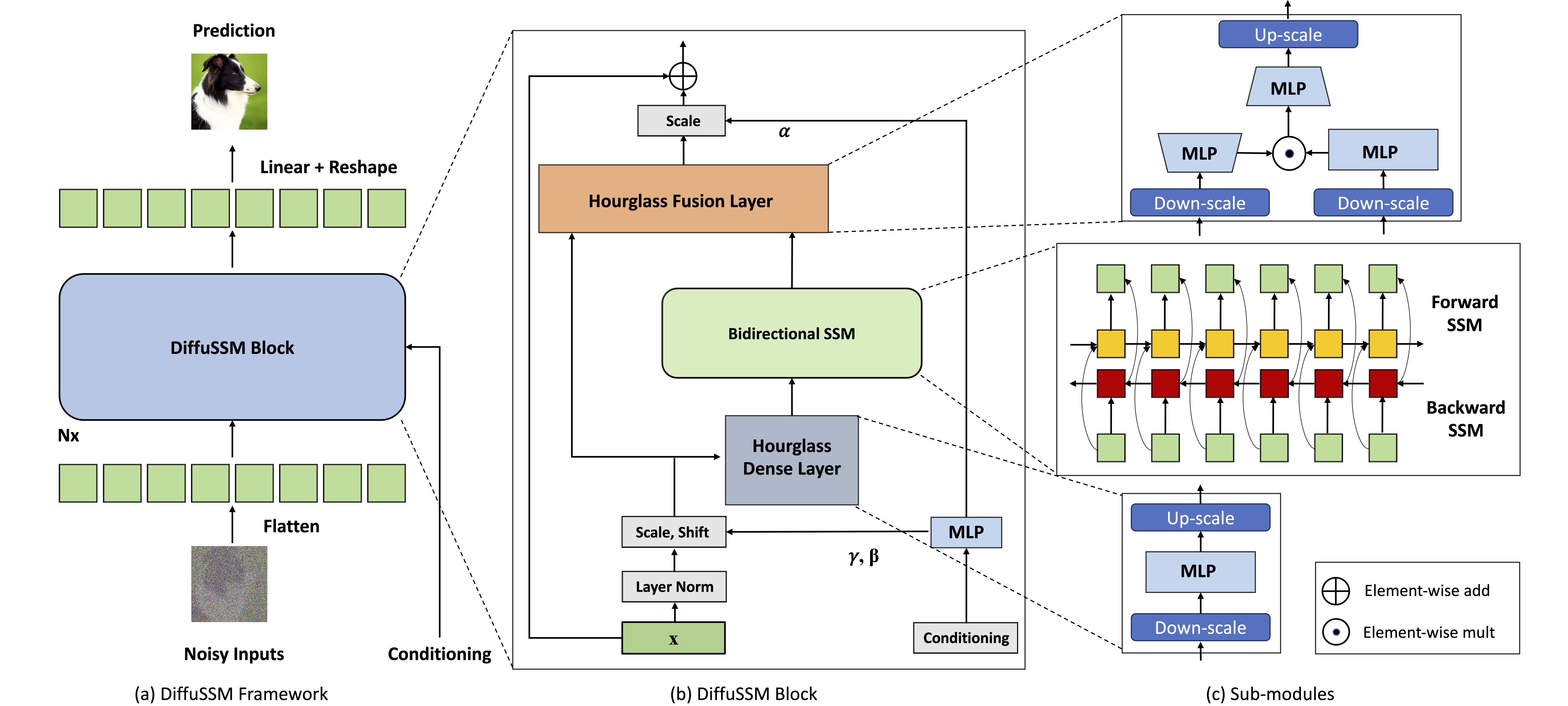}
\caption{Architecture of \model. \model takes a noised image representation which can be a noised latent from a variational encoder, flattens it to a sequence, and applies repeated layers alternating long-range SSM cores with hour-glass feed-forward networks. Unlike with U-Nets or Transformers, there is no application of patchification or scaling for the long-range block.}
\label{fig:dissm:intro:fig}
\end{figure*}


In recent advancements in high-fidelity image generation, Denoising Diffusion Probabilistic Models (DDPMs) have emerged as a key player. However, their application at high resolutions presents significant computational challenges. Current methods, such as patchifying, expedite processes in UNet and Transformer architectures but at the expense of representational capacity. Addressing this, we introduce the Diffusion State Space Model ($\model$), an architecture that supplants attention mechanisms with a more scalable state space model backbone. This approach effectively handles higher resolutions without resorting to global compression, thus preserving detailed image representation throughout the diffusion process. Our focus on FLOP-efficient architectures in diffusion training marks a significant step forward. Comprehensive evaluations on both ImageNet and LSUN datasets at two resolutions demonstrate that DiffuSSMs are on par or even outperform existing diffusion models with attention modules in FID and Inception Score metrics while significantly reducing total FLOP usage. 

\end{abstract}
\pagenumbering{roman}

\section{Introduction}

Rapid progress in image generation has been driven by denoising diffusion probabilistic models (DDPMs) ~\citep{ho2020denoising, nichol2021improved, dhariwal2021diffusion}. DDPMs pose the generative process as iteratively denoising latent variables, yielding high-fidelity samples when enough denoising steps are taken. Their ability to capture complex visual distributions makes DDPMs promising for advancing high-resolution, photorealistic synthesis.

However, significant computational challenges remain in scaling DDPMs to higher resolutions. 
A major bottleneck is the reliance on self-attention~\citep{vaswani2017attention} for high-fidelity generation. 
In U-Nets architectures, this bottleneck comes from combining ResNet~\citep{he2016deep} with attention layers~\citep{van2016conditional, salimans2017pixelcnn++}. DDPMs surpass generative adversarial networks (GANs) but require multi-head attention layers~\citep{nichol2021improved, dhariwal2021diffusion}. 
In Transformer architectures ~\citep{vaswani2017attention}, attention is the central component, and is therefore critical for  achieving recent state-of-the-art image synthesis results~\citep{peebles2022scalable, bao2023all}. 
In both these architectures, the complexity of attention, quadratic in length,  becomes prohibitive when working with high-resolution images. 

Computational costs have motivated the use of representation compression methods. 
 High-resolution architectures generally employ  patchifying~\citep{bao2023all, peebles2022scalable},  or multi-scale resolution\citep{ho2020denoising, nichol2021improved, hoogeboom2023simple}. Patchifying creates coarse-grained representations which reduces computation at the cost of degraded critical high-frequency spatial information and structural integrity~\citep{peebles2022scalable, bao2023all, schramowski2023safe}. Multi-scale resolution, while alleviating computation at attention layers, can diminish spatial details through downsampling~\citep{zamir2021multi} and can introduce artifacts \citep{wang2020deep} while applying up-sampling. 


 The Diffusion State Space Model (\model), is an attention-free diffusion architecture, shown in Figure \ref{fig:dissm:intro:fig}, that aims to circumvent the issues of applying attention for high-resolution image synthesis. \model utilizes a gated state space model (SSM) backbone in the diffusion process. Previous work has shown that sequence models based on SSMs are an effective and efficient general-purpose neural sequence model~\citep{gu2021efficiently}. By using this architecture, we can enable the SSM core to process finer-grained image representations by removing global patchification or multi-scale layers. To further improve efficiency, \model employs an hourglass architecture for the dense components of the network. Together these approaches target the asymptotic complexity of length as well as the practical efficiency in the position-wise portion of the network. 


We validate \model's across different resolutions. Experiments on ImageNet demonstrate consistent improvements in FID, sFID, and Inception Score over existing approaches in various resolutions with fewer total Gflops. 

\section{Related Work}

\paragraph{Diffusion Models}
Denoising Diffusion Probabilistic Models (DDPMs)~\citep{ho2020denoising, nichol2021improved, sohl2015deep, hoogeboom2023simple} are an advancement in the diffusion models family. Previously, Generative Adversarial Networks (GANs)~\citep{goodfellow2014generative} were preferred for generation tasks. Diffusion and score-based generative models~\citep{song2020score,hyvarinen2005estimation, song2019generative, song2023consistency,song2023improved} have shown considerable improvements, especially in image generation tasks~\citep{rombach2022high, saharia2022photorealistic, ramesh2022hierarchical}. Key enhancements in DDPMs have been largely driven by improved sampling methodologies~\citep{ho2020denoising, nichol2021improved, karras2022elucidating}, and the incorporation of classifier-free guidance~\citep{ho2022classifier}.  Additionally, \citet{song2020denoising} has proposed a faster sampling procedure known as Denoising Diffusion Implicit Model(DDIM). Latent space modeling is another core technique in deep generative models. Variational autoencoders (VAEs)~\citep{kingma2013auto} pioneered learning latent spaces with encoder-decoder architectures for reconstruction. 
A similar compression idea was applied in diffusion models as the recent Latent Diffusion Models (LDMs)~\citep{rombach2022high} held state-of-the-art sample quality by training deep generative models to invert a noise corruption process in a latent space when it was first proposed. Additionally, recent approaches also developed masked training procedures, augmenting the denoising training objectives with masked token reconstruction~\citep{gao2023masked, zheng2023fast}. 
Our work is fundamentally built upon existing DDPMs, particularly the classifier-free guidance paradigm. 



\vspace{-10pt}\paragraph{Architectures for Diffusion Models} Early diffusion models utilized U-Net style architectures\citep{ho2020denoising, dhariwal2021diffusion}. Subsequent works enhanced U-Nets with techniques like more layers of attention layers at multi-scale resolution level~\citep{dhariwal2021diffusion, nichol2021improved}, residual connections~\citep{brock2018large}, and normalization~\citep{perez2018film, wu2018group}. However, U-Nets face challenges in scaling to high resolutions due to the growing computational costs of the attention mechanism~\citep{shaham2018spectralnet}. Recently, vision transformers (ViT)~\citep{dosovitskiy2020image} have emerged as an alternate architecture given their strong scaling properties and long-range modeling capabilities proving that convolution inductive bias is not always necessary. Diffusion transformers~\citep{peebles2022scalable, bao2023all} demonstrated promising results. Other hybrid CNN-transformer architectures were proposed~\citep{liu2021swin} to improve training stability. Our work aligns with the exploration of sequence models and related design choices to generate high-quality images but focuses on a complete attention-free architecture.


\vspace{-10pt}\paragraph{Efficient Long Range Sequence Architectures}The standard transformer architecture employs attention to comprehend the interaction of each individual token within a sequence. However, it encounters challenges when modeling extensive sequences due to the quadratic computational requirement. Several attention approximation methods \citep{wang2020linformer, ma2021luna,tay2020sparse, shen2021efficient, hua2022transformer} have been introduced to approximate self-attention within sub-quadratic space. Mega\citep{ma2022mega} combines exponential moving average with a simplified attention unit, surpassing the performance of transformer baselines. Venturing beyond the traditional transformer architectures, researchers are also exploring alternate models adept at handling elongated sequences. State space models (SSM)-based architectures\citep{gu2021efficiently, gupta2022diagonal, gu2022parameterization} have yielded significant advancements over contemporary state-of-the-art methods on the LRA and audio benchmark\citep{goel2022s}. Furthermore, \citet{dao2022hungry, poli2023hyena, peng2023rwkv, qin2023hierarchically} have substantiated the potential of non-attention architectures in attaining commendable performance in language modeling. Our work draws inspiration from this evolving trend of diverting from attention-centric designs and predominantly utilizes the backbone of SSM.



\section{Preliminaries}

\subsection{Diffusion Models}

Denoising Diffusion Probabilistic Model (DDPM)~\citep{ho2020denoising} is a type of generative models that samples images by iteratively denoising a noise input. It starts from a stochastic process where an initial image $x_0$ is gradually corrupted by noise, transforming it into a simpler, noise-dominated state. This forward noising process can be represented as follows:
\begin{align}
&q(x_{1:T} |x_0) = \prod_{t=1}^{T} q(x_t|x_{t-1}), \\
&q(x_t|x_0) = \mathcal{N} (x_t; \sqrt{\bar{\alpha_t}}x_0, (1 - \bar{\alpha_t})I),
\end{align}
where $x_{1:T}$ denotes a sequence of noised images from time $t=1$ to $t=T$.
Then, DDPM learns the \textit{reverse} process that recovers the original image utilizing learned $\mu_{\theta}$ and $\Sigma_{\theta}$: 
\begin{equation}
p_{\theta}(x_{t-1}|x_t) = N (x_{t-1}; \mu_{\theta}(x_t), \Sigma_{\theta}(x_t)),
\end{equation}
where $\theta$ the parameters of the denoiser, and are trained to maximize the variational lower bound~\citep{sohl2015deep} on the log-likelihood of the observed data $x_0$:  $\max_\theta \ -\log p_{\theta}(x_0|x_1) + \sum_t D_{KL}(q^{*}(x_{t-1}|x_t, x_0)\ ||\ p_{\theta}(x_{t-1}|x_t)).$ To simplify the training process, researchers reparameterize $\mu_{\theta}$ as a function of the predicted noise $\varepsilon_{\theta}$ and minimize the mean squared error between $\varepsilon_{\theta}(x_t)$ and the true Gaussian noise $\varepsilon_t$: $\min_{\theta} ||\varepsilon_{\theta}(x_t) - \varepsilon_t||^2_2.$
However, to train a diffusion model that can learn a variable reverse process covariance $\Sigma_{\theta}$, we need to optimize the full $L$. In this work, we follow DiT~\cite{peebles2022scalable} to train the network where we use the simple objective to train the noise prediction network $\varepsilon_{\theta}$ and use the full objective to train the covariance prediction network $\Sigma_{\theta}$.
After training is done, we follow the stochastic sampling process to generate images from the learned  $\varepsilon_{\theta}$ and $\Sigma_{\theta}$.

\subsection{Architectures for Diffusion Models}


We review methods for 
parameterizing $\mu_\theta$ which maps $\mathbb{R}^{H \times W \times C} \rightarrow \mathbb{R}^{H \times W \times C}$ 
where $H, W, C$ are the height, width, and size of the data. For image generation tasks, they can either raw pixels, or some latent space representations extracted from a pre-trained VAE encoder~\citep{rombach2022high}. When generating high-resolution images, even in the latent space, $H$ and $W$ are large, and require specialized architectures for this function to be tractable. 
\vspace{-10pt}\paragraph{U-Nets with Self-attention}
U-Net architectures~\citep{ho2020denoising, nichol2021improved, hoogeboom2023simple} uses
convolutions and sub-sampling at multiple resolutions to handle high-resolution inputs, where additional self-attention layers are used at each low-resolution blocks.
To the best of our knowledge, no U-Net-based diffusion models are achieving state-of-the-art performance without using self-attention. Let $t_1, \ldots t_T$ be a series of lower-resolution feature maps created by down-sampling the image.\footnote{Note that choices of up- and down-scale include learned parameters and non-parameterized ones such as average pooling and upscale~\citep{hoogeboom2023simple,croitoru2023diffusion}.} At each scale a ResNet~\citep{he2016deep} is applied to $\mathbb{R}^{H_{t} \times W_{t} \times C_{t}}$. These are then upsampled and combined into the final output. 
To enhance the performance of U-Net in image generation, attention layers are integrated at the lowest-resolutions. The feature map is flattened to a sequence of $H_{t} W_{t}$ vectors. For instance, when considering  $H=256\times W=256$ down to attention layers of $16\times 16$ and $32\times 32$, leading to sequences of length $256$ and $1024$ respectively. Applying attention earlier improves accuracy at a larger computational cost.
More recently, \citep{hoogeboom2023simple,podell2023sdxl} have shown that using more self-attention layers in the low-resolution is the key of scaling high-resolution U-Net-based diffusion models.



\vspace{-10pt}\paragraph{Transformers with Patchification} 
As mentioned above, the global contextualization using self-attention is the key for diffusion models to perform well. Therefore, it is also natrual to consider architecture fully based on self-attention.
Transformer architectures utilize attention throughout, but handle high-resolution images through patchification~\citep{dosovitskiy2020image}. Given a patch size $P$, the transformer partitions the image into $P \times P$ patches yielding a new $\mathbb{R}^{H/P \times W/P \times C'}$ representation. 
This patch size $P$ directly influences the effective granularity of the image and downstream computational demands. To feed patches into a Transformer, the image is flattened and a linear embedding layer is applied to obtain a sequence of $(HW)/P^2$ hidden vectors~\citep{peebles2022scalable, bao2023all, hoogeboom2023simple, dosovitskiy2020image}. Due to this embedding step, which projects from $C'$ to the model size, large patches risk loss of spatial details and ineffectively model local relationships due to reduced overlap. However, patchification has the benefit of reducing the quadratic cost of attention as well as the feed-forward networks in the Transformer.

\section{{\model}}
Our goal is to design a diffusion architecture that learns long-range interactions at high-resolution without requiring ``length reduction'' like patchification. Similar to DiT, the approach works by flattening the image and treating it like a sequence modeling problem. However, unlike Transformers, this approach uses sub-quadratic computation in the length of this sequence.


\subsection{State Space Models (SSMs)}

SSMs are a class of architectures for processing discrete-time sequences~\citep{gu2021efficiently}. The models behave like a linear recurrent neural network (RNN) processing an input sequence of scalars $u_1, \ldots u_L$ to output $y_1, \ldots y_L$ with the following equation,
\begin{align*}
x_{k} = \boldsymbol{\overline{A}} x_{k-1} + \boldsymbol{\overline{B}} u_k,  \ \ \ 
y_k = \boldsymbol{\overline{C}} x_k. 
\end{align*}
Where $\boldsymbol{\overline{A}} \in \mathbb{R}^{N\times N}, \boldsymbol{\overline{B}} \in \mathbb{R}^{N\times 1},\boldsymbol{\overline{C}} \in \mathbb{R}^{1\times N}$.
The main benefit of this approach, compared to alternative architectures such as Transformers and standard RNNs, is that the linear structure allows it to be implemented using a long \textit{convolution} as opposed to a recurrence. Specifically, $y$  can be computed from $u$ with an FFT yielding $O(L\log L)$ complexity, allowing it to be applied to significantly longer sequences. When handling vector inputs, we can stack $D$ different SSMs and apply a $D$ batched FFTs. 

However a linear RNN, by itself, is not an effective sequence model. 
The key insight from past work is that if the discrete-time values $\boldsymbol{\overline{A}}, \boldsymbol{\overline{B}},\boldsymbol{\overline{C}}$ are derived from appropriate continuous-time 
state-space models, the linear RNN approach can be made stable and effective~\citep{gu2020hippo}. We therefore learn a continuous-time SSM parameterization $\boldsymbol{A},\boldsymbol{B},\boldsymbol{C}$ as well as a discretization rate $\Delta$, which is used to produce the necessary discrete-time parameters. Original versions of this conversion were challenging to implement, however recently researchers~\citep{gu2022parameterization, gupta2022diagonal} have introduced simplified diagonalized versions of SSM neural networks that achieve comparable results with a simple approximation of the continuous-time parameterization.  We use one of these, S4D~\citep{gu2022parameterization}, as our backbone model. 

Just as with standard RNNs, SSMs can be made bidirectional by concatenating the outputs of two SSM layers and passing them through an MLP to yield a $L \times 2D$ output. In addition, past work shows that this layer can be combined with multiplicative gating to produce an improved Bidirectional SSM layer~\citep{wang2022pretraining, mehta2022long} as part of the encoder, which is the motivation for our architecture.

\subsection{\model Block}
The central component of our \model is a gated bidirectional SSM, aimed at optimizing the handling of long sequences. To enhance efficiency, we incorporate \emph{hourglss} architectures within MLP layers. This design alternates between expanding and contracting sequence lengths around the Bidirectional SSMs, while specifically reducing sequence length in MLPs. The complete model architecture is shown in Figure~\ref{fig:dissm:intro:fig}.

Specifically, each hourglass layer receives a shortened, flattened input sequence $\mathbf{I} \in \mathbb{R}^{J \times D}$ where $M = L / J$ is the downscale and upscale ratio. At the same time, the entire block including the bidirectional SSMs is computed in the original length to fully leverage the global contexts. We use $\sigma$ to denote activation functions. We compute the following for $l \in \{1\ldots L\}$ with $j = \lfloor l / M \rfloor, m = l\ \text{mod}\ M, D_m = 2D / M$. 
\begin{align*}
& \mathbf{U}_l =\sigma (\mathbf{W}^{\uparrow}_{k} \sigma(\mathbf{W}^0 \mathbf{I}_j)     &&  \in \mathbb{R}^{L\times D}\\
& \mathbf{Y} = \text{Bidirectional-SSM}(\mathbf{U}) && \in \mathbb{R}^{L\times 2D}\\
& \mathbf{I}'_{j, D_mk:D_m(k+1)}  = \sigma(\mathbf{W}^{\downarrow}_{k} \mathbf{Y}_l) && \in \mathbb{R}^{J\times 2D}\\
& \mathbf{O}_j= \mathbf{W}^3 (\sigma(\mathbf{W}^2\mathbf{I}'_j) \odot \sigma(\mathbf{W}^1 \mathbf{I}_j))     && \in \mathbb{R}^{J\times D} 
\end{align*}
We integrate this Gated SSM block in each layer with a skip connection. Additionally, following past work we integrate a combination of the class label $\mathbf{y} \in \mathbb{R}^{L \times 1}$ and timestep $\mathbf{t} \in \mathbb{R}^{L \times 1}$ at each position, as illustrated in Figure~\ref{fig:dissm:intro:fig}. 

\begin{figure}[t]\vspace{-2em}
    \centering    
    \includegraphics[width=1.1\linewidth]{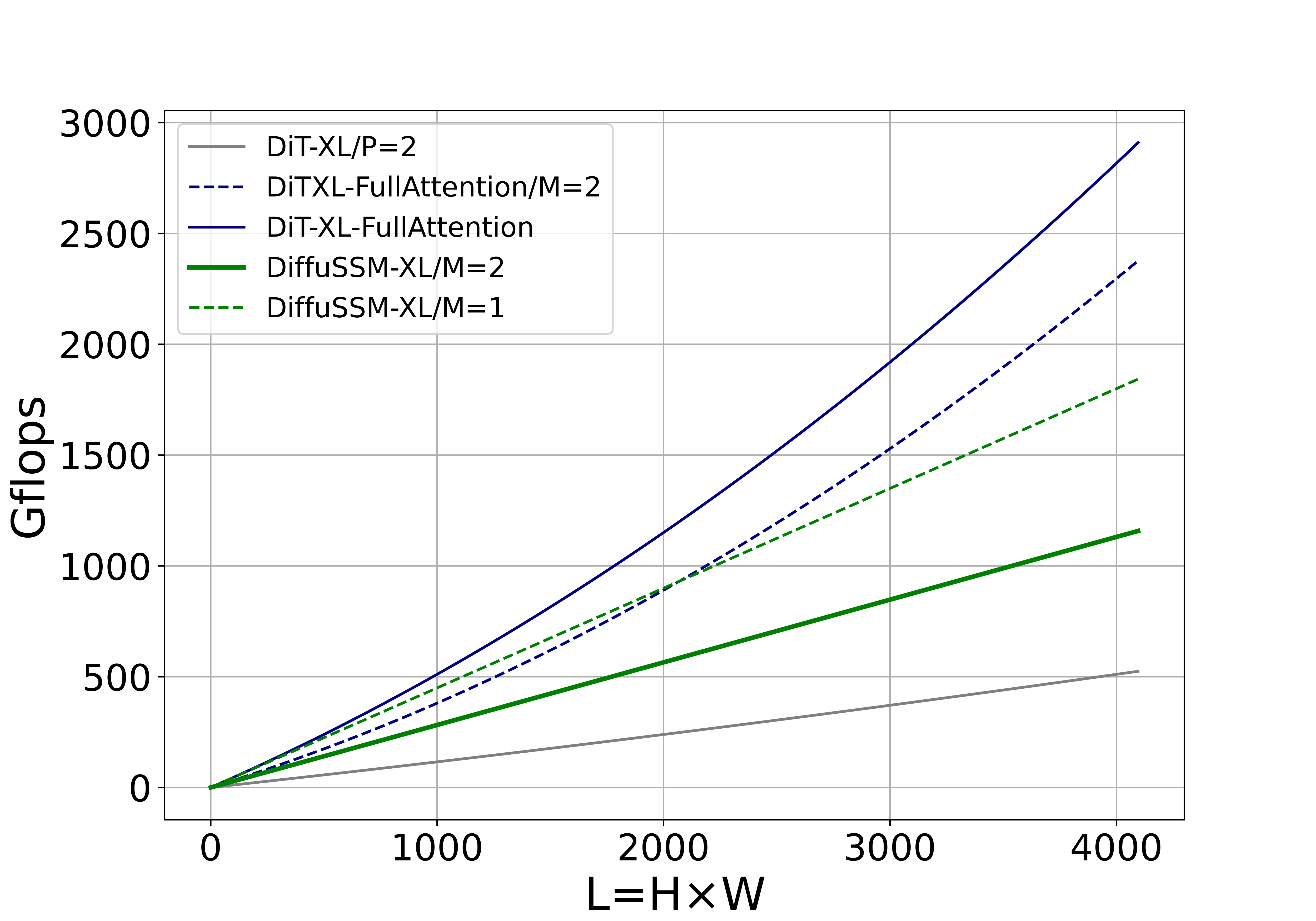}
\caption{Comparison of Gflops of DiT and \model under various model architecture. DiT with patching (P=2) scales  well to longer sequences, however when patching is removed it scales poorly even with hourglass (M=2). \model scales well, and hourglass (M=2) can be used to reduce absolute Gflops.}\label{fig:flops}
\end{figure}

\paragraph{Parameters} 
The number of parameters in the \model block is dominated by the linear transforms, $\boldsymbol{W}$, these contain $9D^2 + 2MD^2$ parameters. With $M=2$ this yields $13D^2$ parameters. The DiT transformer block has $12D^2$ parameters in its core transformer layer; however, the DiT architecture has more parameters in other layer components (adaptive layer norm). We match parameters in experiments by using an additional \model layer.

\vspace{-1em}\paragraph{FLOPs}
Figure~\ref{fig:flops} compares the Gflops between DiT and \model.
The total Flops in one layer of \model 
is $13 \frac{L}{M} D^2 + LD^2 + \alpha 2\ L \log L D$ where $\alpha$ represents a constant for the FFT implementation. With $M=2$ and noting that the linear layers dominate computation, this yields roughly $7.5LD^2$ Gflops. In comparison, if instead of using SSM, we had used self-attention at full length with this hourglass architecture, we would have $2DL^2$ additional Flops. 


Considering our two experimental scenarios: 1) $D\approx L=1024$ which would have given $2LD^2$ extra Flops, 2)  $4D\approx L=4096$ which would give $8LD^2$ Flops and significantly increase cost.  As the core cost at Bidirectional SSM is small compared to that using attention, and as a result using hourglass architecture will not work for attention-based models.  DiT avoids these issues by using patching as discussed earlier, at the cost of representational compression.

\section{Experimental Studies}
\subsection{Experimental Setup}
\begin{figure*}[t]
    \centering
    \includegraphics[width=\linewidth]{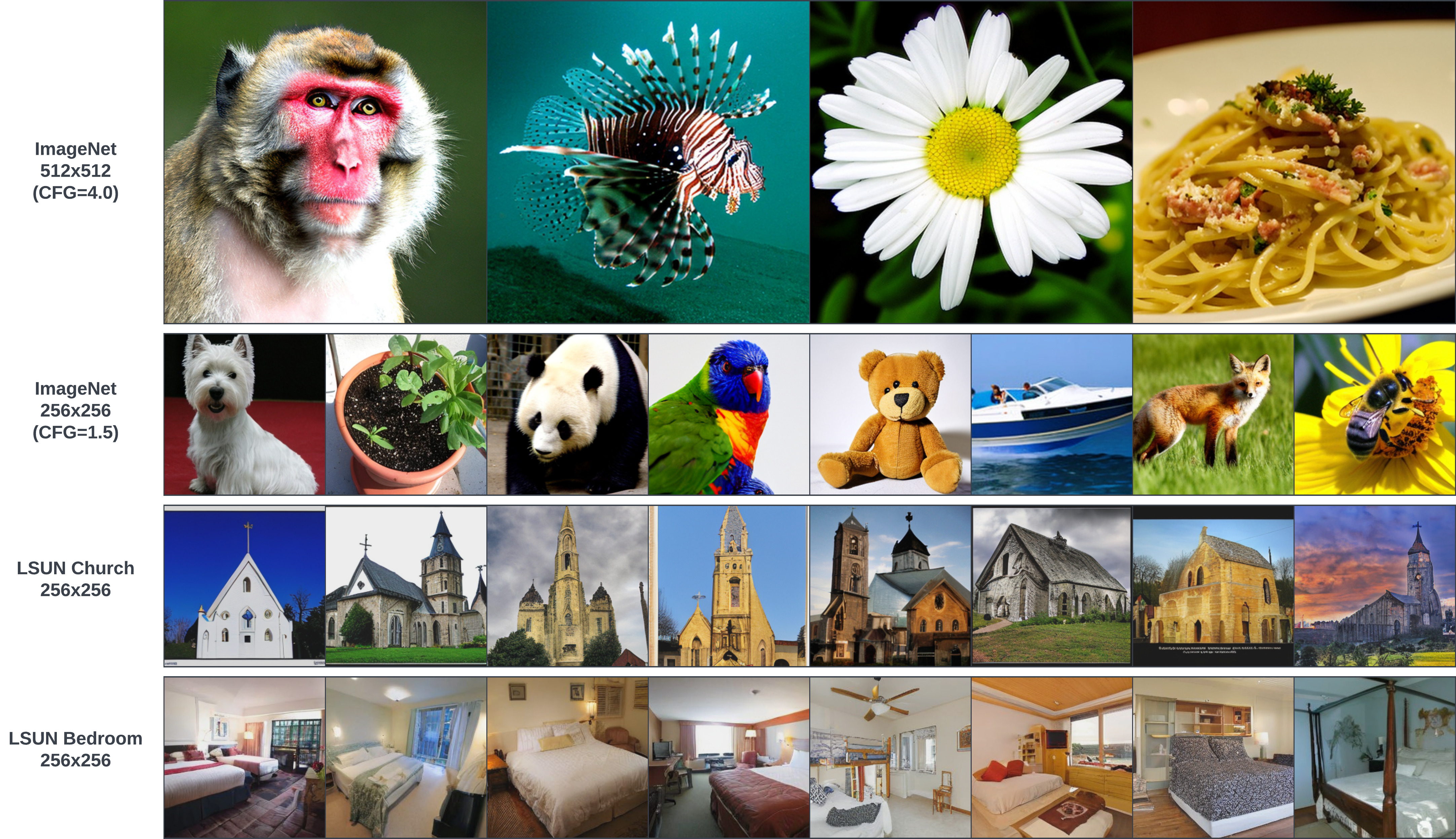}
\caption{Uncurated samples from the {\model} models trained from various datasets.}
    \label{fig:enter-label}
\end{figure*}

\begin{table*}[t]
\centering
\begin{tabular}{l|cc|ccccc}
\midrule
\midrule

\multicolumn{5}{l}{\textbf{ImageNet 256×256 Benchmark}} \\
\midrule
     Models  & Total & Total  & FID $\downarrow$ & sFID $\downarrow$& IS $\uparrow$& P $\uparrow$ & R $\uparrow$ \\
     & Images(M) & Gflops  &  &  &  &  &  \\
\midrule
BigGAN-deep~\citep{brock2018large} & -   & - & 6.95 &   7.36 &     171.40  &  0.87 &   0.28               \\
MaskGIT~\citep{chang2022maskgit} & 355  & -  &6.18  &  -  &  182.1  &  0.80 &   0.51               \\
StyleGAN-XL~\citep{sauer2022stylegan} & - & - &  \textbf{2.30}  &  4.02    & 265.12 &  0.78 &  0.53      \\
\midrule
ADM~\citep{dhariwal2021diffusion}  & 507 &  $5.68\times10^{12}$ &  10.94  &  6.02 &   100.98   &  0.69 &     0.63   \\
ADM-U~\citep{dhariwal2021diffusion}  & 507 &   $3.76\times10^{11}$  &     7.49  & 5.13 &   127.49   &  0.72 &     0.63   \\
\arrayrulecolor{mylightgray} 
\hline
CDM~\citep{ho2022cascaded} & - & - & 4.88  &   -  & 158.71  &   -  &     -       \\
\arrayrulecolor{mylightgray} 
\hline
\arrayrulecolor{mylightgray} 
\hline
LDM-8~\citep{rombach2022high}  & 307 & $1.75\times10^{10}$ & 15.51   &  - &   79.03    & 0.65  &  0.63        \\
LDM-4~\citep{rombach2022high} &  213 & $2.22\times10^{10}$ &  10.56   &  - &    103.49    & 0.71  &  0.62        \\
DiT-XL/2~\citep{peebles2022scalable} & 1792 &  $2.13\times10^{11}$  & 9.62  &  6.85   &  121.50    & 0.67 &   0.67       \\
\arrayrulecolor{black}
\textbf{\model-XL}  & 660 & $1.85\times10^{11}$ &  \textbf{9.07}   &  5.52 &   118.32   & 0.69    &  0.64       \\
\midrule
\midrule
Classifier-free Guidance \\
\midrule
ADM-G  &  507 &  $5.68\times10^{11}$  & 4.59  &   5.25   & 186.70 & 0.82   &  0.52    \\
ADM-G, ADM-U  & 507 & $3.76\times10^{12}$ &  3.60  &   -   & 247.67 & 0.87   & 0.48       \\
LDM-8-G  & 307 & $1.75\times10^{10}$   &    7.76  & -   & 209.52    & 0.84 &  0.35    \\
LDM-4-G  &  213 &  $2.22\times10^{10}$  &  3.95 &   -   &   178.2 2 & 0.81   &  0.55      \\
U-ViT-H/2-G~\citep{bao2023all}  &  512  & $6.81\times10^{10} $& 2.29  &   -   & 247.67 & 0.87   & 0.48       \\
DiT-XL/2-G  & 1792 &   $2.13\times10^{11}$  &  \textbf{2.27} &   4.60   & \textbf{278.24}  & 0.83  &  0.57    \\
\textbf{\model-XL}-G &  660 &  $1.85\times10^{11}$  & 2.28 &  \textbf{4.49}     & 259.13  & 0.86  &  0.56    \\
\midrule
\midrule
\multicolumn{5}{l}{\textbf{ImageNet 512×512 Benchmark}} \\
\midrule
ADM &  1385  & $5.97\times 10^{11}$  & 23.24 &   10.19  & 58.06  &   0.73    &  0.60        \\
ADM-U & 1385 &  $3.9\times 10^{12}$  & 9.96 &   5.62   & 121.78  &   0.75   &  0.64       \\
ADM-G & 1385 & $5.97\times 10^{11}$   & 7.72    & 6.57   &  172.71    & 0.87  &   0.42    \\
ADM-G, ADM-U & 1385 &  $4.5\times 10^{12}$ &  3.85 &  5.86    & 221.72  & 0.84 &   0.53      \\
\arrayrulecolor{mylightgray} 
\hline
U-ViT/2-G  & 512 &  $6.81\times 10^{10}$  &   4.05   & 8.44 &  \textbf{261.13} &  0.84     & 0.48   \\
DiT-XL/2-G   & 768 & $4.03\times 10^{11}$ & \textbf{3.04} &   \textbf{5.02}   &  240.82 & 0.84  &0.54          \\
\arrayrulecolor{black} 
\textbf{\model-XL-G}  & 302  & $3.22\times 10^{11}$ &  3.41    &  5.84 & 255.06  & 0.85 & 0.49        \\
\hline
\end{tabular}
\caption{Class conditional image generation quality evaluation of \model and existing approaches
on ImageNet 256$\times$ 256. Reported results from other cited papers with their $\#$ trained images. Total images by training steps $\times$ batch size as reported, and total Gflops by Total Images $\times$ GFlops/Image. P refers to Precision and R refers to Recall. $-G$ denotes
the results with classifier-free guidance. }\label{tab:imagenet}
\end{table*}

\begin{table}
\begin{tabular}{l|cc|cc}
\midrule
Models& \multicolumn{2}{c}{LSUN-Church}&  \multicolumn{2}{c}{LSUN-Bedroom}\\
 & FID$\downarrow$& P $\uparrow$&  FID$\downarrow$&P $\uparrow$\\
\midrule
ImageBART~\citep{esser2021imagebart}& 7.32& &  5.51&-\\
PGGAN~\citep{karras2017progressive}& 6.42& -&  -&-\\
StyleGAN~\citep{karras2019style}& 4.21& -& 2.35& 0.59\\
StyleGAN2~\citep{karras2020analyzing}& 3.93& 0.39& -& -\\
ProjGAN~\citep{sauer2021projected}& \textbf{1.59}& 0.61& 1.52& 0.61\\
\midrule
DDPM~\citep{ho2020denoising} & 7.89 & - & 4.90 & - \\
UDM~\citep{kim2021soft}& -& -& 4.57&-\\
ADM~\citep{dhariwal2021diffusion} & -& -& \textbf{1.90}&\textbf{0.66}\\
LDM~\citep{rombach2022high}& 4.02& \textbf{0.64}& 2.95& \textbf{0.66}\\
\model & 3.94& \textbf{0.64}& 3.02 & 0.62 \\\hline
\end{tabular}    
\caption{Unconditional image generation evaluation of \model and exsiting approaches on LSUN-Church and LSUN-Bedroom at $256\times 256$. }\label{tab:lsun}
\end{table}

\paragraph{Datasets} Our primary experiments are conducted on  ImageNet\citep{deng2009imagenet}\footnote{https://image-net.org/download.php} and LSUN\citep{yu2015lsun}\footnote{https://www.yf.io/p/lsun}. Specifically, we used the ImageNet-1k dataset where there are $~1.28$ million images and $1000$ classes of objects. For the LSUN-dataset, we choose two categories: Church (126k images) and Bed (3M images), and train separate unconditional models for them.  Our experiments are conducted with the ImageNet dataset at $256\times256$ and $512\times512$ resolution, and LSUN at $256\times256$ resolution.  We use latent space encoding\cite{rombach2022high} which gives effective sizes $32\times 32$ and $64\times 64$ with $L = 1024$ and $L=4096$ respectively.  We also include pixel-space ImageNet at $128\times128$ resolution in our supplementary materials where $L=16,384$.




\paragraph{Linear Decoding and Weight Initialization} 
After the final block of the Gated SSM, the model decodes the sequential image representation to the original spatial dimensions to output noise prediction and diagonal covariance prediction. Similar to \citet{peebles2022scalable, gao2023masked}, we use a linear decoder and then rearrange the representations to obtain the original dimensionality. We follow DiT to use the standard layer initializations approach from ViT~\citep{dosovitskiy2020image}.

\vspace{-10pt}\paragraph{Training Configuration} We followed the same training recipe from DiT~\citep{peebles2022scalable} to maintain an identical setting across all models. We also chose to follow existing literature to keep an exponential moving average (EMA) of model weights with a constant decay. 
Off-the-shelf VAE encoders from \footnote{https://github.com/CompVis/stable-diffusion} were used, with parameters fixed during training. 
 Our \model-XL possesses approximately $673$M parameters and encompasses 29 layers of Bidirectional Gated SSM blocks with a model size $D=1152$. This value is similar to DiT-XL.  trained our model using a mixed-precision training approach to mitigate computational costs.
We adhere to the identical configuration of diffusion as outlined in ADM~\citep{dhariwal2021diffusion}, including their linear variance scheduling, time and class label embeddings, as well as their parameterization of covariance $\Sigma_{\theta}$. More details can be found in the Appendix. 

For unconditional image generation, DiT does not report results and we were unable to compare with DiT in the same training setting. Our objective instead compares \model, with a training regimen comparable to taht of LDM\cite{rombach2022high} that can generate high-quality images for categories in the LSUN dataset. To adapt the model to an unconditional context, we have removed the class label embedding. 



\paragraph{Metrics} To quantify the performance of image generation of our model, we used Frechet Inception Distance(FID)~\citep{heusel2017gans}, a common metric measuring the quality of generated images. We followed convention when comparing against prior works and reported FID-50K using 250 DDPM sampling steps. We also reported sFID score~\citep{nash2021generating}, which is designed to be more robust to spatial distortions in the generated images. For a more comprehensive insight, we also presented the Inception Score~\citep{salimans2016improved} and Precision/Recall~\citep{kynkaanniemi2019improved} as supplementary metrics. Note that do not incorporate classifier-free guidance unless explicitly mentioned(we used $-G$ for the usage of classifier-free guidance or explicitly state the CFG).  

\vspace{-10pt}\paragraph{Implementation and Hardware} We implemented all models in Pytorch and trained
them using NVIDIA A100. \model-XL, our most compute-intensive model trains on 8 A100 GPUs 80GB with a global batch size of 256. More computation details and speed can be found in the supplementary materials. 

\subsection{Baselines}
We compare to a set of previous best models, these include: GAN-style approaches that previously achieved state-of-the-art results, UNet-architectures trained with pixel space representations, and Transformers operating in the latent space. More details can be found in Table \ref{tab:image:net:256}. Our aim is to compare, through a similar denoising process, the performance of our model with respect to other baselines. Some recent studies~\citep{zheng2023fast, gao2023masked} focusing on image generation at the $256 \times 256$ resolution level have combined masked token prediction with existing DDPM training objectives to advance the state of the art. However, these works are orthogonal to our primary comparison, so we have not included them in Table 1. For LSUN datasets, we found existing DDPM-based methods are not surpassing GAN-based methods. Our goal is to compare within the DDPM framework instead of competing with state-of-the-art methods.



\subsection{Experimental Results}\label{tab:image:net:256}
\paragraph{Class-Conditional Image Generation}
We compare \model with state-of-the-art class-conditional generative models, as depicted in Table \ref{tab:imagenet}. When classifier-free guidance is not employed, \model outperforms other diffusion models in both FID and sFID, reducing the best score from the previous non-classifier-free latent diffusion models from $9.62$ to $9.07$, while utilizing $\sim3\times$ fewer training steps. In terms of Total Gflops of training, our uncompressed model yields a ~$20\%$ reduction of the total Gflops compared with DiT. When classifier-free guidance is incorporated, our models attain the best sFID score among all DDPM-based models, exceeding other state-of-the-art strategies, demonstrating the images generated by \model are more robust to spatial distortion. As for FID score, \model surpasses all models when using classifier-free guidance, and maintains a pretty small gap ($0.01$) against DiT.  Note that \model trained with $30\%$ fewer total Gflops already surpasses DiT when no classifier-free guidance is applied. U-ViT~\citep{bao2023all} is another transformer-based architecture but uses a UNet-based architecture with long-skip connections between blocks. U-ViT used fewer FLOPs and yielded better performance at a 256\(\times\)256 resolution, but this is not the case for the 512\(\times\)512 dataset. As our major comparison is against DiT, we do not adopt this long-skip connection for a fair comparison. We acknowledge that adapting U-Vit's idea might benefit both DiT and \model. We leave this consideration for future work.

We further compare on a higher-resolution benchmark using classifier-free guidance. Results from \model here are relatively strong and near some of the state-of-the-art high-resolution models, beating all models but DiT on sFID and achieving comparable FID scores.  The \model was trained on 302M images, seeing $40\%$ as many images and using $25\%$ fewer Gflops as DiT.  

\begin{figure*}[tbh]
    \centering    
    \includegraphics[width=1\linewidth]{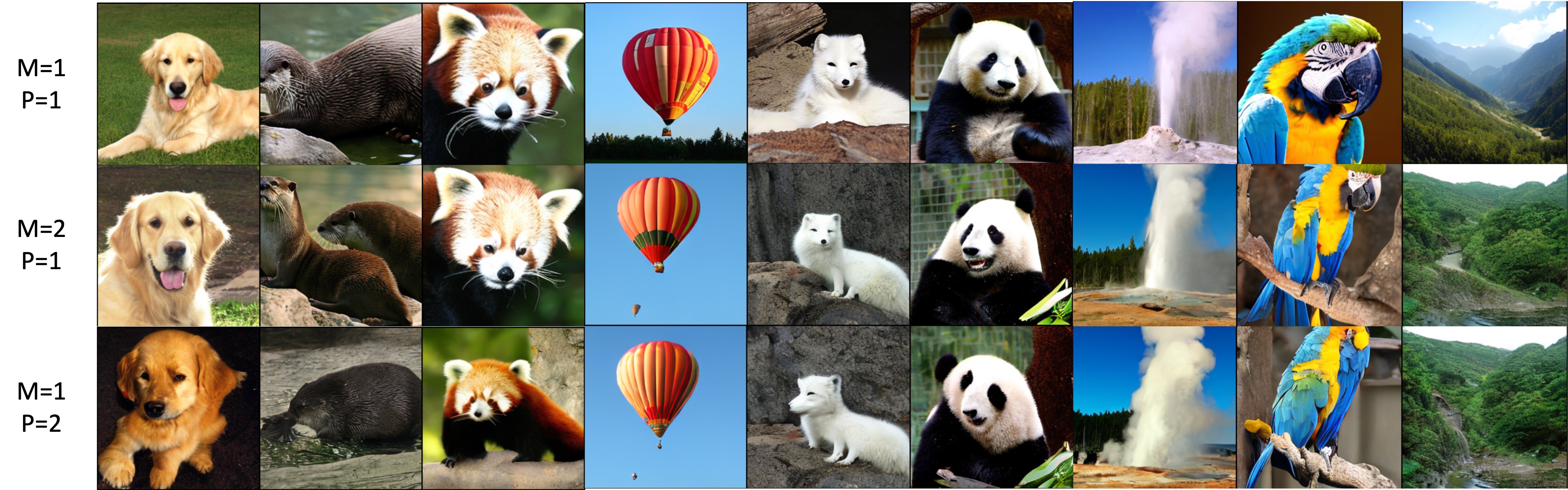}
\caption{Qualititative studies of patching and down/up scale of \model. $P$ refers to the patchfication, $M$ refers to the down/up scale ratio. $P=1$ is the case where there is not patchfication and $M=1$ is the case where there is no down/up scale.  }
\label{fig:qual}
\end{figure*}
\begin{figure}[tbh]
    \centering    
    \includegraphics[width=1\linewidth]{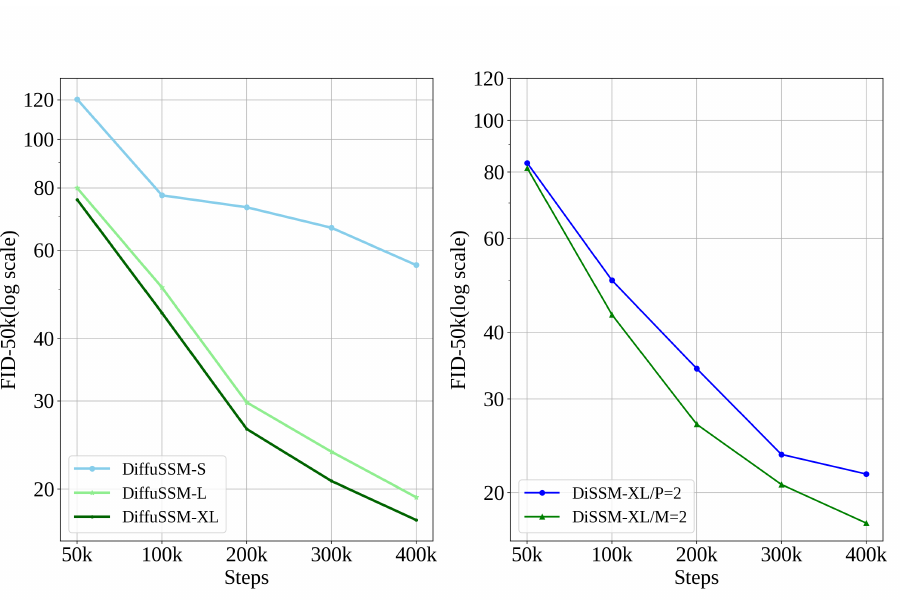}
    \caption{Ablation studies. Left: \model with different hidden dimension sizes D($-S/D=384, -L/D=786, -XL/D=1152$). Right: FID score of \model with different patch size ($P=2$) and downsample ratio ($M=1$).}\label{fig:ablation}
\end{figure}

\paragraph{Unconditional Image Generation} We compare the unconditional image generation ability of our model against existing baselines. Results are shown in Table \ref{tab:lsun}. Our findings indicate that \model achieves comparable FID scores obtained by LDM (with $-0.08$ and $0.07$ gap) with a comparable training budget. This result highlights the applicability of \model across different benchmarks and different tasks. Similar to LDM, our approach doesn't outperform ADM for LSUN-Bedrooms as we are only using $~25\%$ of the total training budget as ADM.  For this task, the best GAN models outperform diffusion as a model class.


\vspace{-1em}\section{Analysis}

\begin{figure*}[t]\vspace{-1em}
    \centering
    \includegraphics[width=1\linewidth]{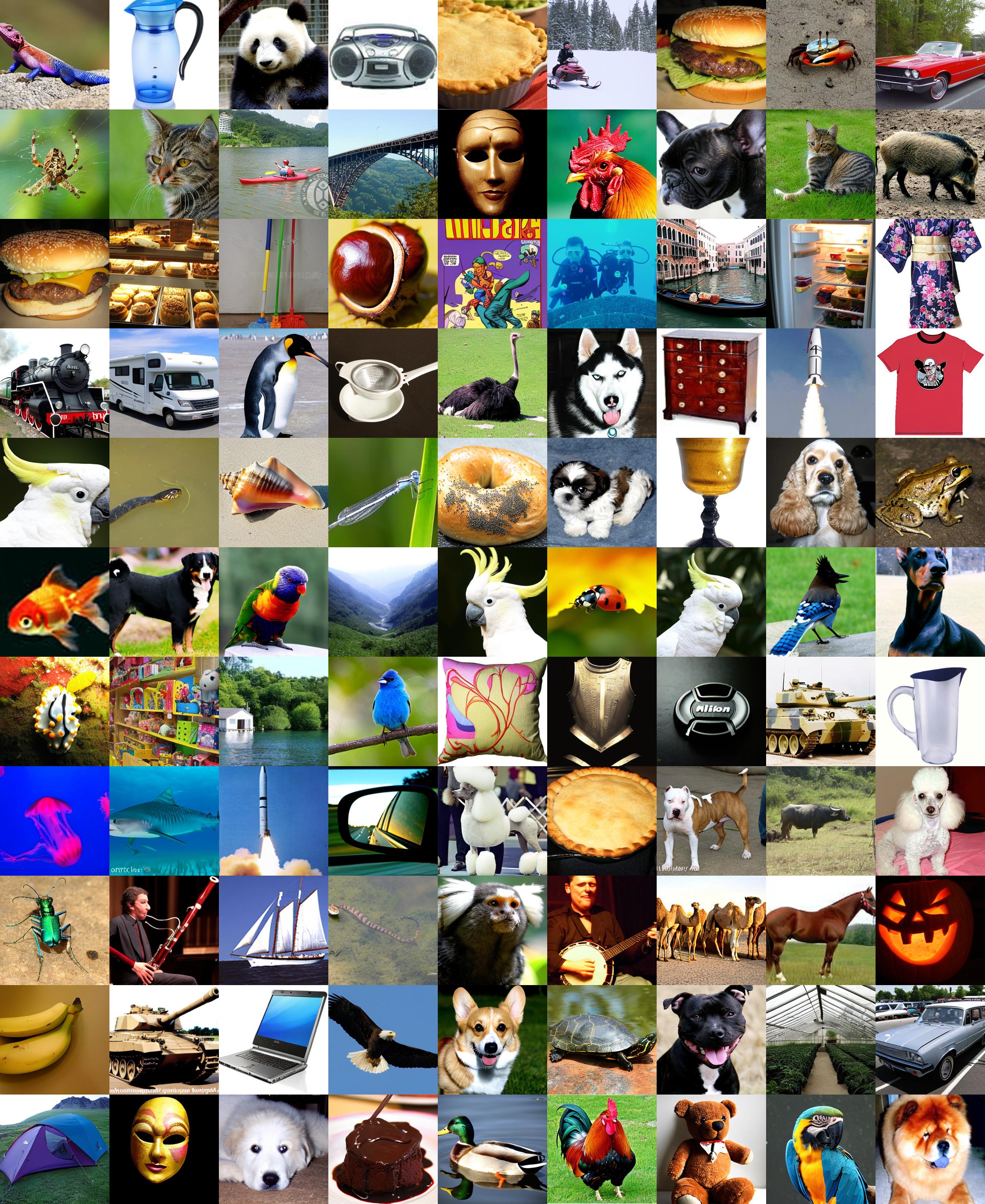}
    \caption{Samples from the {\model} models on ImageNet 256 $\times$ 256.}
    \label{fig:addtional1}
\end{figure*}

\begin{figure*}[t]\vspace{-1em}
    \centering
    \includegraphics[width=1\linewidth]{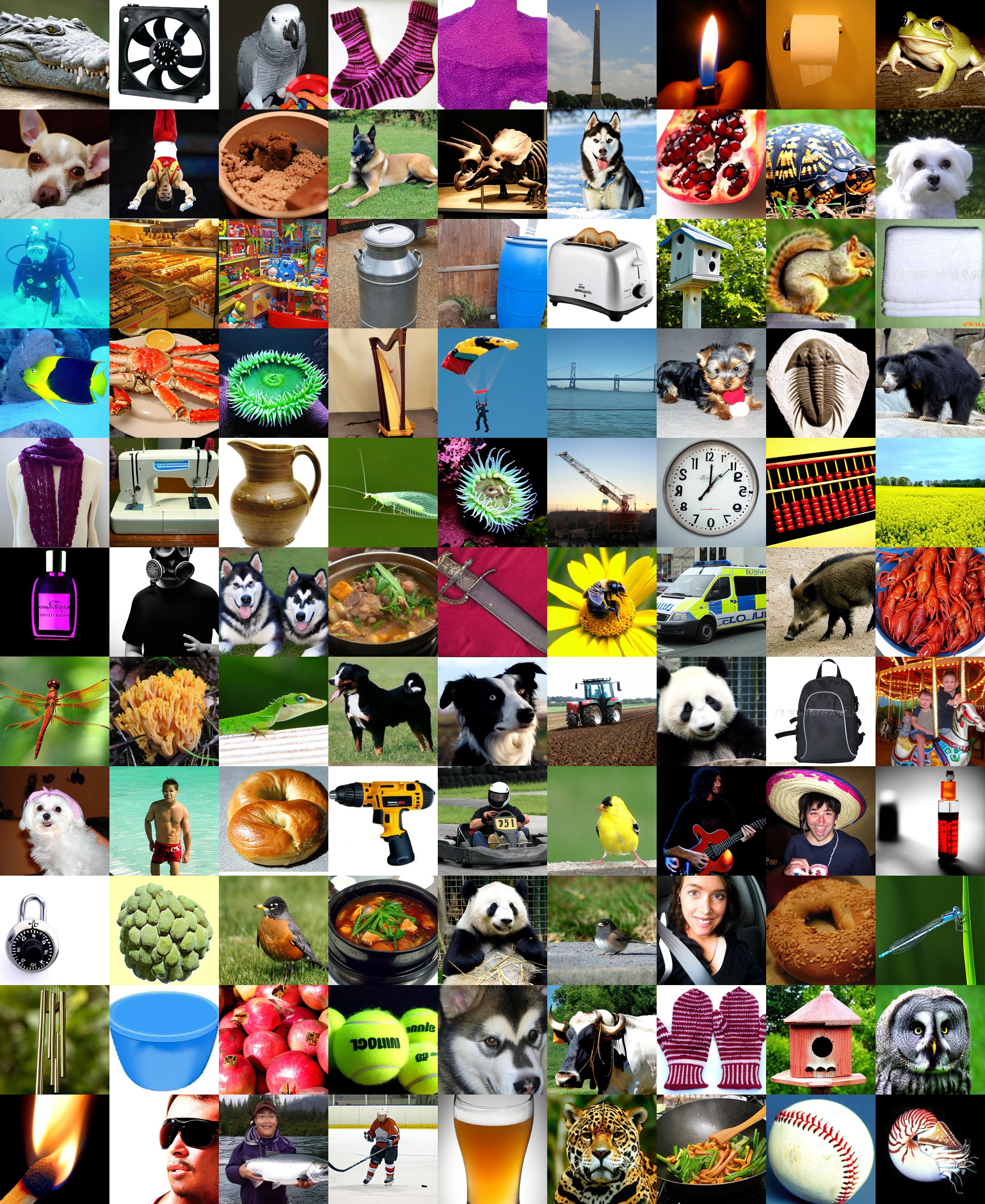}
    \caption{Samples from the {\model} models on ImageNet 256 $\times$ 256.}
    \label{fig:enter-label6}
\end{figure*}

\begin{figure*}[t]\vspace{-1em}
    \centering
    \includegraphics[width=1\linewidth]{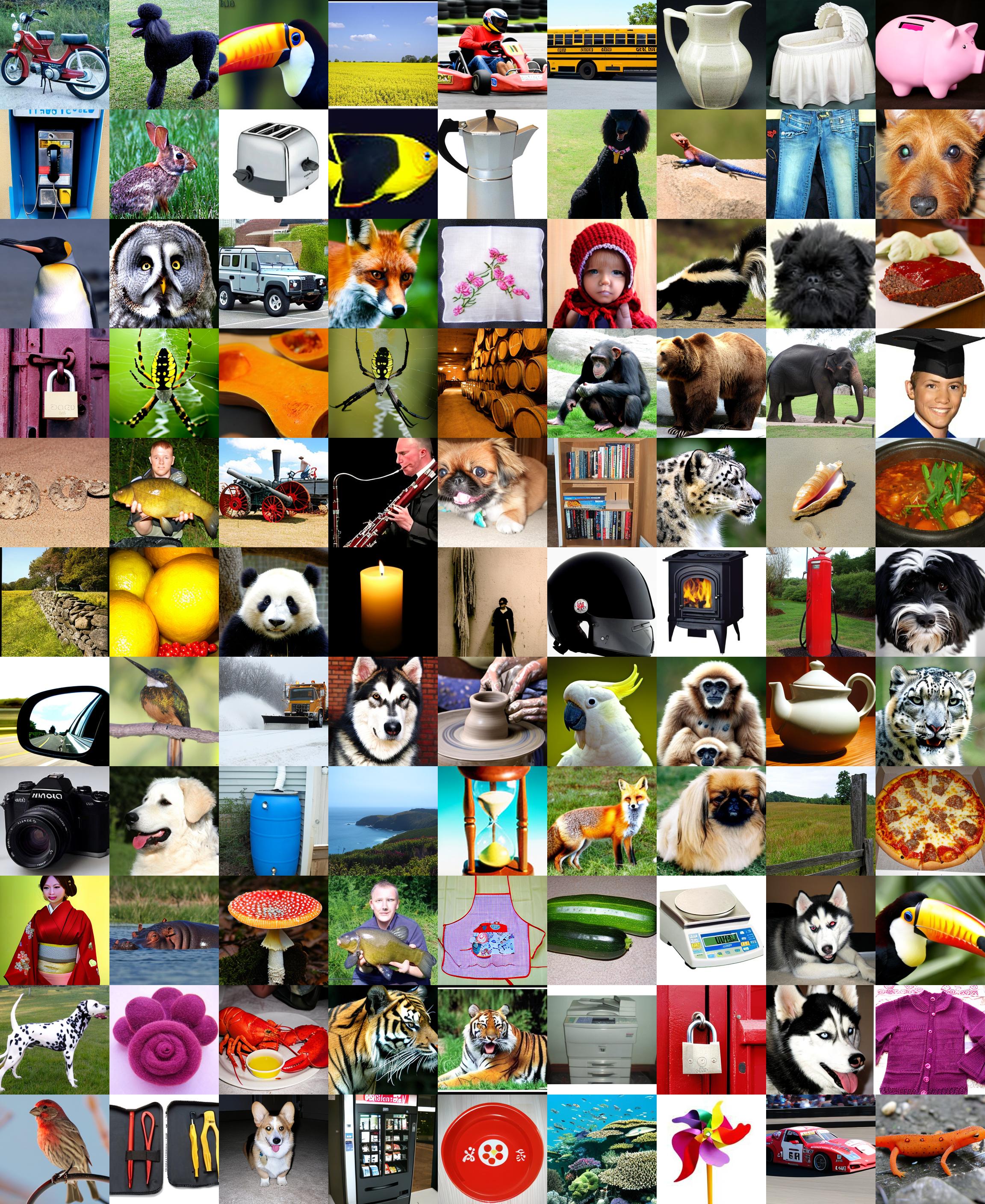}
    \caption{Samples from the {\model} models on ImageNet 256 $\times$ 256.}
    \label{fig:enter-label5}
\end{figure*}

\begin{figure*}[t]\vspace{-1em}
    \centering
    \includegraphics[width=1\linewidth]{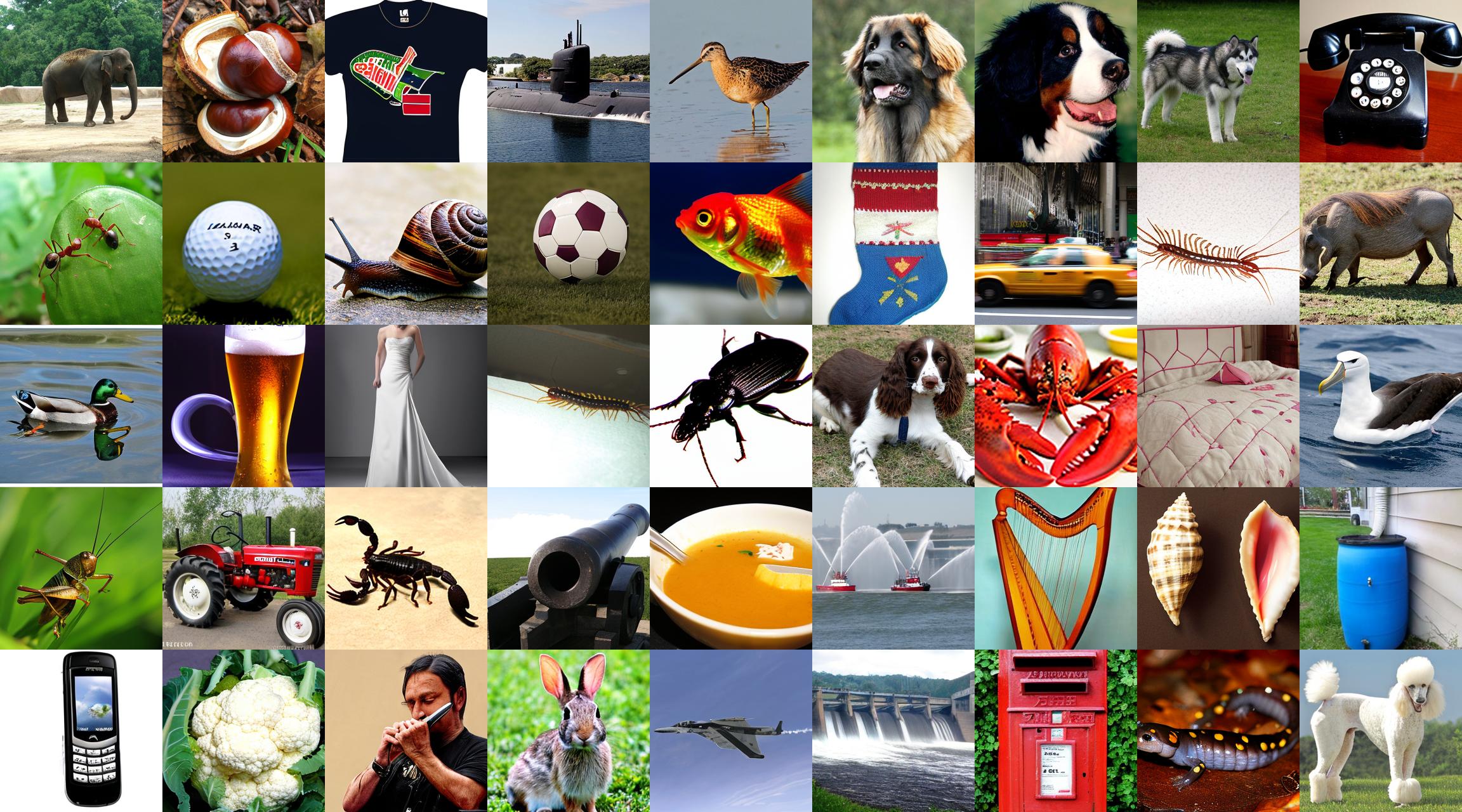}
    \caption{Samples from the {\model} models on ImageNet 256 $\times$ 256.}
    \label{fig:enter-label4}
\end{figure*}

\begin{figure*}[t]\vspace{-1em}
    \centering
    \includegraphics[width=.92\linewidth]{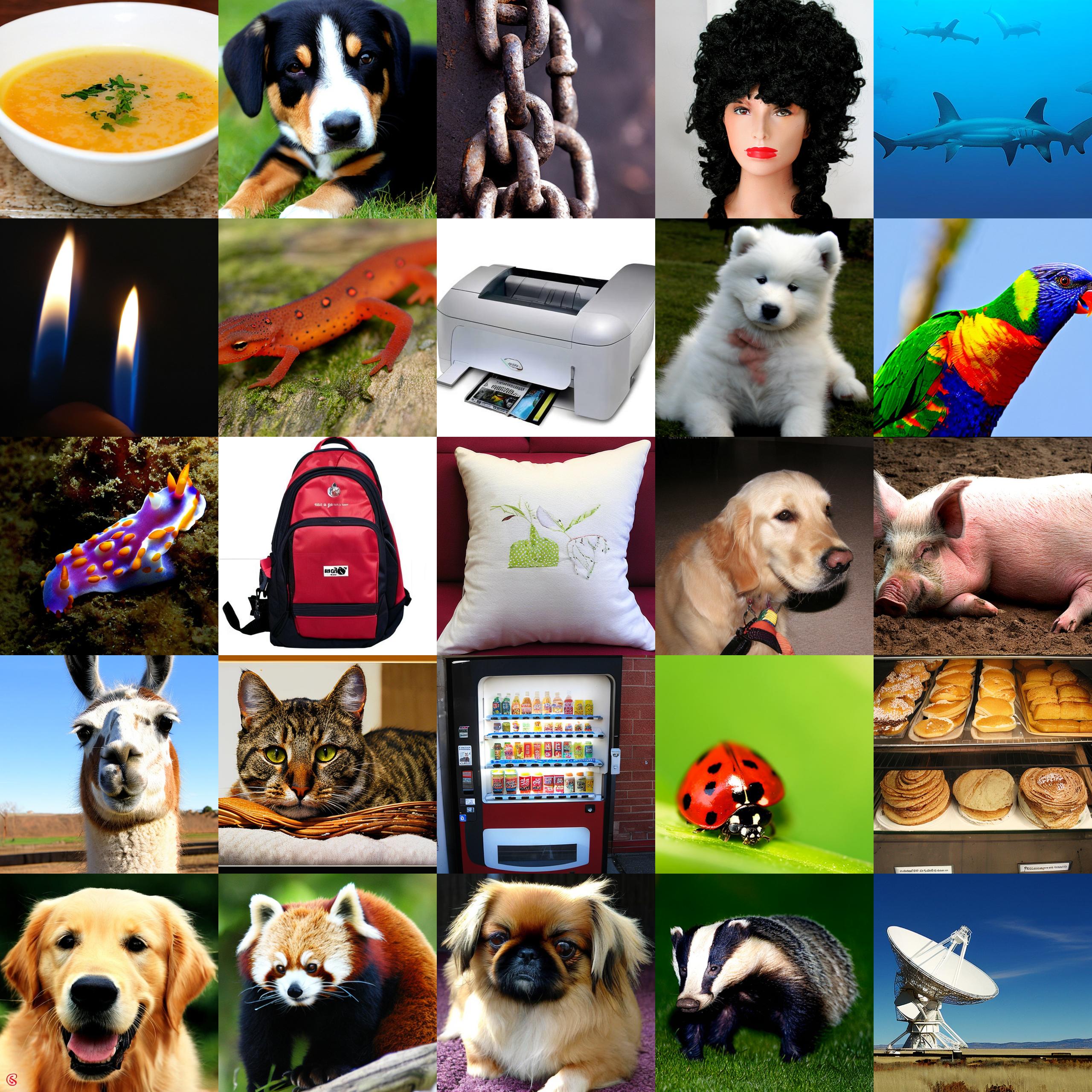}
    \caption{Samples from the {\model} models on ImageNet 512 $\times$ 512.}
    \label{fig:enter-label3}
\end{figure*}

\begin{figure*}[t]\vspace{-1em}
    \centering
    \includegraphics[width=0.92\linewidth]{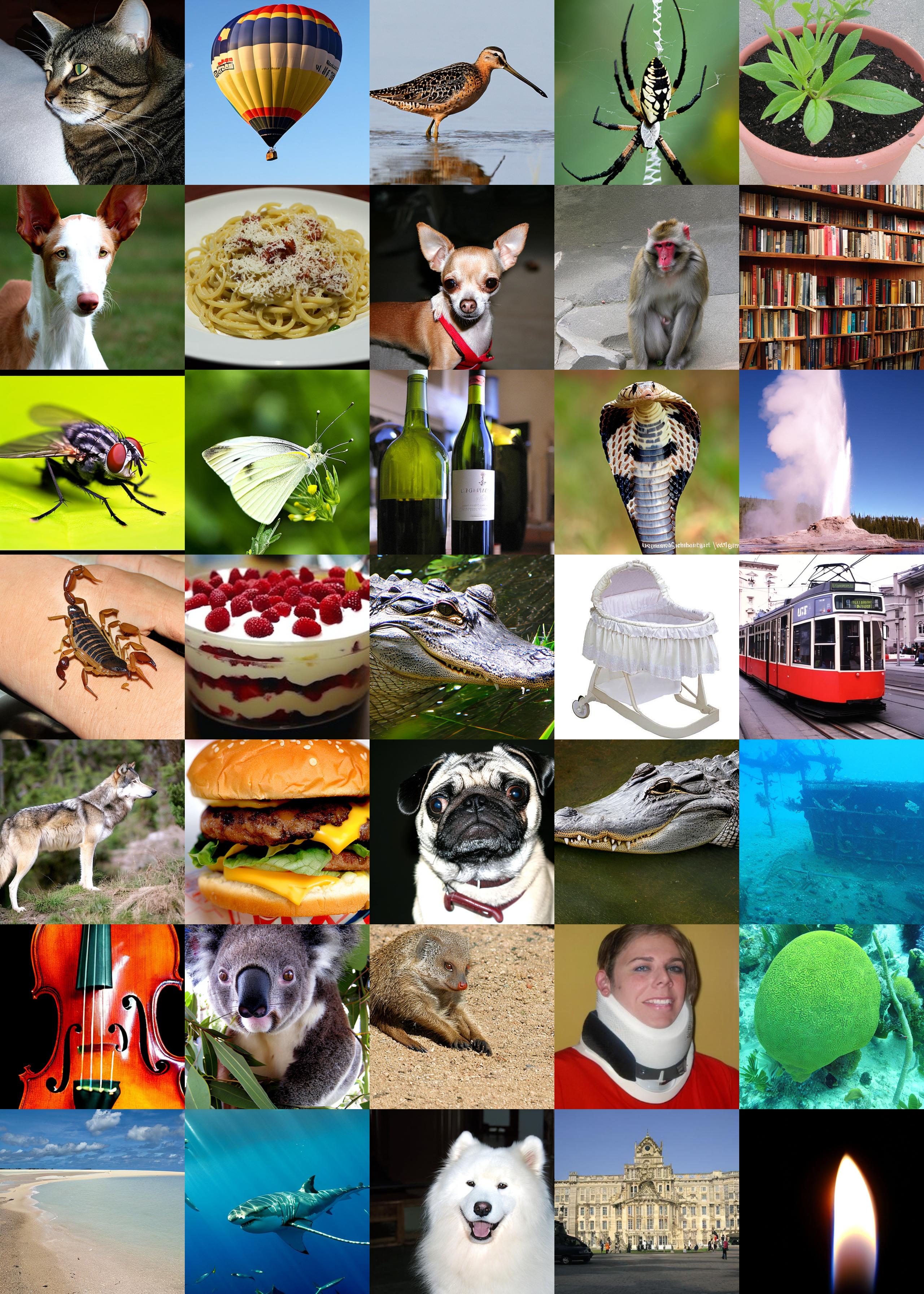}
    \caption{Samples from the {\model} models on ImageNet 512 $\times$ 512.}
    \label{fig:enter-label2}
\end{figure*}

\begin{figure*}[t]\vspace{-1em}
    \centering
    \includegraphics[width=0.92\linewidth]{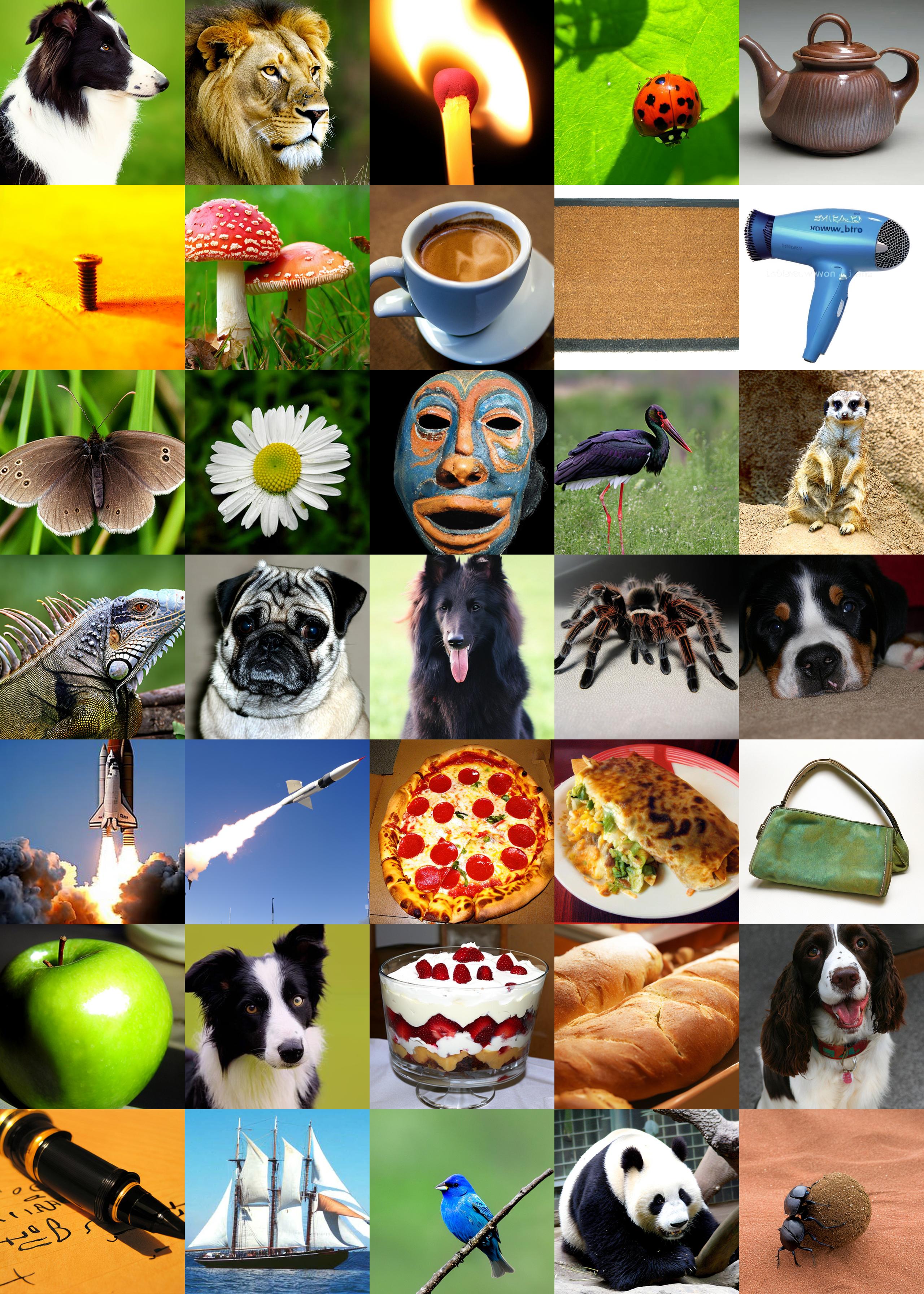}
    \caption{Samples from the {\model} models on ImageNet 512 $\times$ 512.}
    \label{fig:enter-label1}
\end{figure*}

\begin{figure*}[t]\vspace{-1em}
    \centering
    \includegraphics[width=0.92\linewidth]{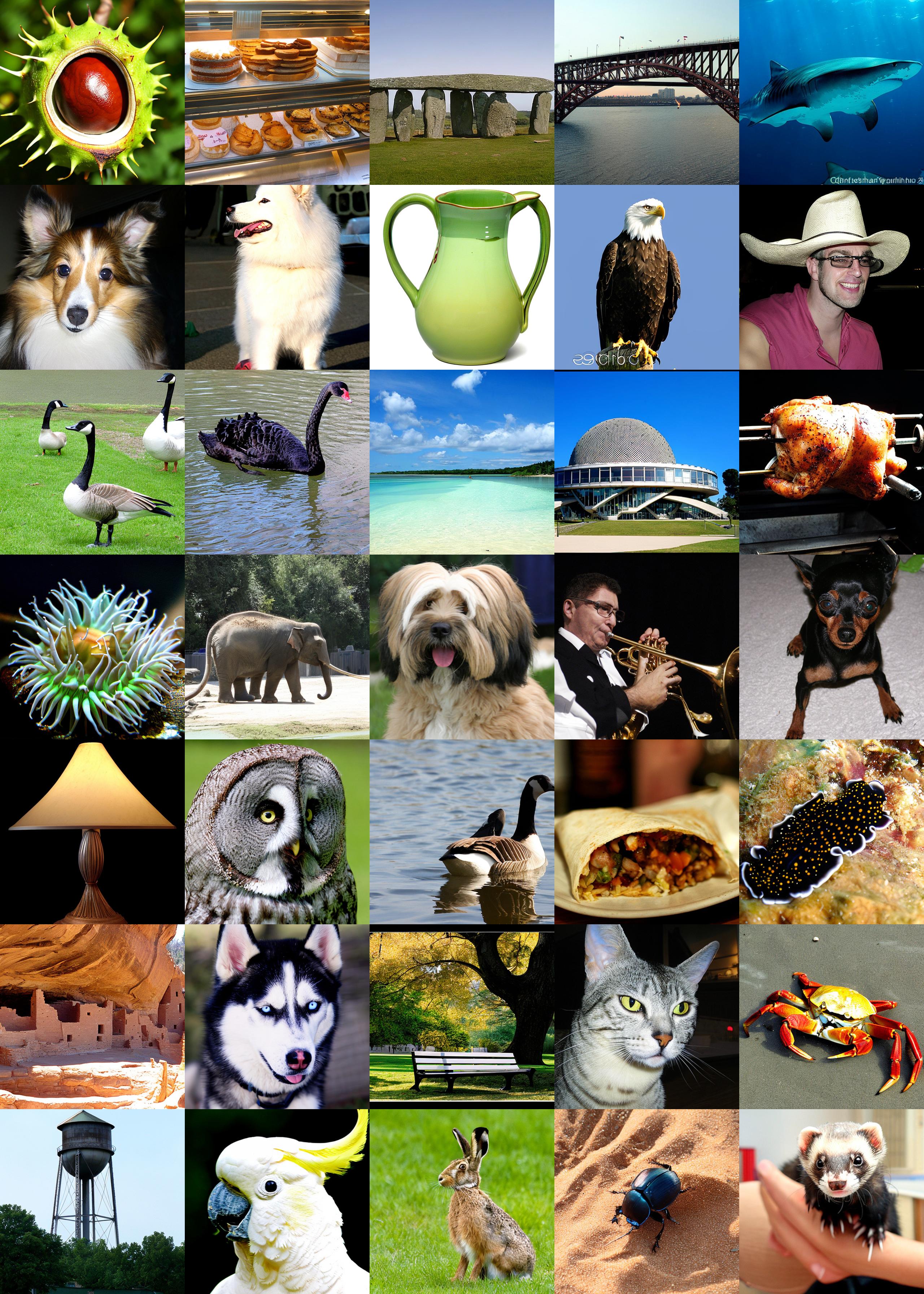}
    \caption{Samples from the {\model} models on ImageNet 512 $\times$ 512.}
    \label{fig:addtional1:last}
\end{figure*}

\paragraph{Additional Images}Additional images generated by \model are included from Figure \ref{fig:addtional1} to Figure \ref{fig:addtional1:last}. 

\paragraph{Model Scaling}
We trained three different \model sizes to calibrate the performance yielded by scaling up the model. We calculate the FID-50k for their checkpoints of the first 400k steps. Results are shown in Figure~\ref{fig:ablation}  (Left). We find that similar to DiT models, large models use FLOPs more efficient and scaling the \model will improve the FID at all stages of training.

\paragraph{Impact of Hourglass}
We trained our model with different sampling settings to assess the impact of compression in latent space: using a downsampling ratio \(M=2\) (our regular model), and another with \(P=2\) applying a patch size equal to 2, similar to what DiT has done. We calculated their FID-50k for the first 400k steps and plotted it on a log scale. Results are shown in Figure~\ref{fig:ablation} (Right).  We find that our model yields a better FID score compared to when patching is applied, and the gap between the two also widens as the number of training steps increases. This suggests that the compression of information might hurt the model's ability of generating high-quality images.



\paragraph{Qualitative Analysis} The objective of \model is to avoid compressing hidden representations. To test whether this is beneficial we compare three variants of \model with different downscale ratio $M$ and patch size $P$. We train all three model variants for 400K steps with the same batch size and other hyperparameters. When generating images, we use identical initial noise and noise schedules across class labels. Results are presented in Figure~\ref{fig:qual}. Notably, eliminating patching enhances robustness in spatial reconstruction at the same training stages. This results in improved visual quality, comparable to uncompressed models, but with reduced computation.


\section{Conclusion}

We introduce \model, an architecture for diffusion models that does not require the use of Attention. This approach can handle long-ranged hidden states without requiring representation compression. Results show that this architecture can achieve better performance than DiT models utilizing less Gflops at 256x256 and competitive results at higher-resolution even with less training. 
The work has a few remaining limitations. First, it focuses on (un)conditional image generation as opposed to full text-to-image approaches. Additionally, there are some recent approaches such as masked image training that may improve the model.  Still this model provides an alternative approach for learning effective diffusion models at large scale. 
We believe removing the attention bottleneck should open up the possibility of applications in other areas that requires long-range diffusion, for example high-fidelity audio, video, or 3D modeling.

{
    \small
    \bibliographystyle{ieeenat_fullname}
    \bibliography{ref}
}


\end{document}